\def\year{2022}\relax
\documentclass[letterpaper]{article} % DO NOT CHANGE THIS
\usepackage{aaai22}  % DO NOT CHANGE THIS
\usepackage{times}  % DO NOT CHANGE THIS
\usepackage{helvet}  % DO NOT CHANGE THIS
\usepackage{courier}  % DO NOT CHANGE THIS
\usepackage[hyphens]{url}  % DO NOT CHANGE THIS
\usepackage{graphicx} % DO NOT CHANGE THIS
\usepackage{natbib}  % DO NOT CHANGE THIS AND DO NOT ADD ANY OPTIONS TO IT
\usepackage{caption} % DO NOT CHANGE THIS AND DO NOT ADD ANY OPTIONS TO IT
\DeclareCaptionStyle{ruled}{labelfont=normalfont,labelsep=colon,strut=off} % DO NOT CHANGE THIS
\usepackage{algorithm}
\usepackage{algorithmic}
\usepackage{lineno} % add line number
\usepackage{amsmath}
\usepackage{amsfonts}
\usepackage{booktabs}
\usepackage{makecell}
\usepackage{hyperref}
\usepackage{url}
\usepackage{changes}
\usepackage{subfigure}
\usepackage{afterpage}
\usepackage{enumitem}
\usepackage{multirow}
\usepackage{booktabs}
\usepackage{xcolor}
\newcommand{\answerYes}[1]{\textcolor{blue}{#1}} 
\newcommand{\answerNo}[1]{\textcolor{teal}{#1}} 
\newcommand{\answerNA}[1]{\textcolor{gray}{#1}}

\usepackage{newfloat}
\usepackage{listings}
\floatstyle{ruled}
\newfloat{listing}{tb}{lst}{}
\floatname{listing}{Listing}
\title{Paths of A Million People: Extracting Life Trajectories from Wikipedia}

\author{
    Ying Zhang\equalcontrib, Xiaofeng Li\equalcontrib, Zhaoyang Liu, Haipeng Zhang\thanks{Corresponding author.}
}
\affiliations{
    ShanghaiTech University\\
    \{zhangying12022, lixf2022, liuzhy2023, zhanghp\}@shanghaitech.edu.cn
}

\usepackage{bibentry}
\begin{document}
% \linenumbers % add line number
\maketitle

\begin{abstract}
% editor 建议把第一句话拆成两句
% Notable people's life trajectories have been a focus of study -- the locations and times of various activities, such as birth, death, education, marriage, competition, work, delivering a speech, making a scientific discovery, finishing a masterpiece, and fighting a battle, and how these people interact with others, carry important messages for the broad research related to human dynamics.
The life trajectories of notable people have been studied to pinpoint the times and places of significant events such as birth, death, education, marriage, competition, work, speeches, scientific discoveries, artistic achievements, and battles. Understanding how these individuals interact with others provides valuable insights for broader research into human dynamics.
However, the scarcity of trajectory data in terms of volume, density, and inter-person interactions, limits relevant studies from being comprehensive and interactive. We mine millions of biography pages from Wikipedia and tackle the generalization problem stemming from the variety and heterogeneity of the trajectory descriptions. Our ensemble model \textbf{COSMOS}, which combines the idea of semi-supervised learning and contrastive learning, achieves an F1 score of $\textbf{85.95\%}$. For this task, we also create a hand-curated dataset, \textit{WikiLifeTrajectory}, consisting of 8,852 \textit{(person, time, location)} triplets as ground truth. Besides, we perform an empirical analysis on the trajectories of 8,272 historians to demonstrate the validity of the extracted results. To facilitate the research on trajectory extractions and help the analytical studies to construct grand narratives, we make our code, the million-level extracted trajectories, and the \textit{WikiLifeTrajectory} dataset publicly available\footnote{\url{https://github.com/ZhangDataLab/COSMOS} Currently, we publish 5\% extracted data. The remaining data and the \textit{WikiLifeTrajectory} dataset will be released with the camera-ready version.}.

\end{abstract}

\section{Introduction}

\label{intro}

Life trajectories~\cite{elder1994time} of notable individuals have profound and crucial implications. These people choose where to study, live, work, campaign, and spend the rest of their lives~\cite{elder1994time,doherty2009potus,schich2014network}, while engaging in social interactions~\cite{Becker1974ATO}. Throughout these processes, new ideas emerge and spread~\cite{elder2003emergence}, communities are established, clusters are formed~\cite{schatzki2022trajectories}, and technologies are created~\cite{schatzki2019social}. \citet{schich2014network} analyze the birth and death places of more than 150,000 notable people and reveal how cultural centers evolve over a span of more than 2,000 years. Studies also pay attention to scientists' relocation, and discover the shifts in scientists' career choices~\cite{deville2014career} and the power comparisons between nations~\cite{verginer2020cities}, and how these moves affect scientific productions and progresses~\cite{zucker2009star}. Additionally, the whereabouts of politicians imply political influences~\cite{goldsmith2021does}, alliances~\cite{doherty2009potus}, and regional socioeconomic developments~\cite{boianovsky20182017}.

However, the trajectory data is scarce in terms of volume, density, and inter-person interactions, limiting previous studies from being comprehensive and interactive. For instance, the aforementioned cultural history study~\cite{schich2014network} does not go beyond birth and death places, while the places people choose to study and work in their prime, may better shape cultural centers. Besides, with these intermediate points, the life trajectories can be ``densified'' to tell more complete stories. Furthermore, if social interactions were available, the life trajectories would literally intersect, transforming into much richer dynamic spatio-temporal networks of people.
Many other studies that rely on data of small scales~\cite{Morrissey2015ArchivesOC,doherty2009potus}, can hardly be extended to grand narratives across time and space. Their data sources, such as hand-curate databases~\cite{Hautala2019artist}, government websites~\cite{doherty2009potus,goldsmith2021does}, and freebase~\cite{schich2014network}, can hardly provide the desirable dataset, and this is the reason that we turn to the entire Wikipedia. According to our statistics, the raw Wikipedia dumps\footnote{We use the Wikipedia dumps of June 1, 2023, from \url{https://dumps.wikimedia.org/}.} contain 1,930,519 biography pages and on average each page has 11 locations and 15 time entities, serving as a convenient source of life trajectories.

\begin{figure}[!htbp]
\centering
\subfigure[]{
\begin{minipage}[!htb]{1\linewidth}
\centering % ,height=0.8\linewidth
\includegraphics[width=1\textwidth]{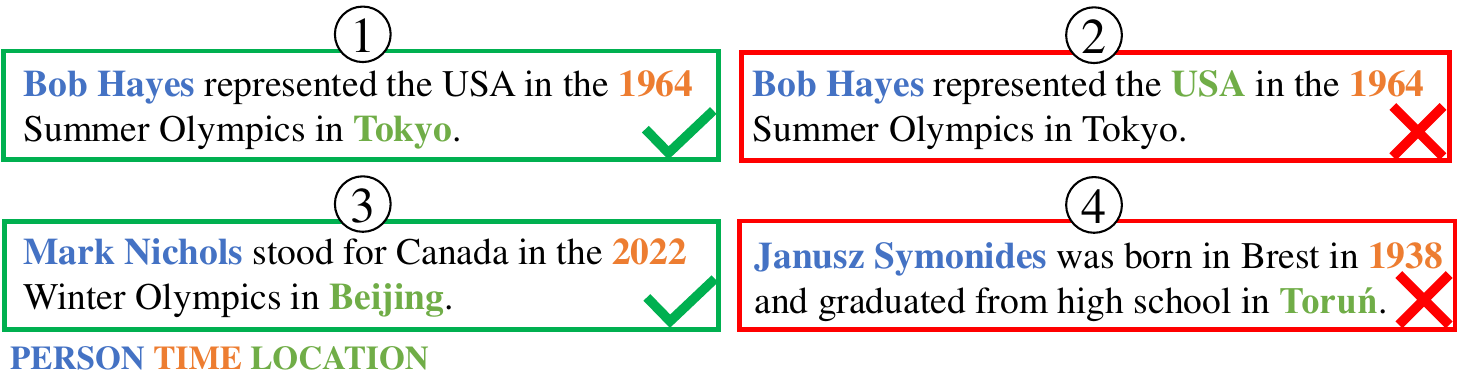}
\label{fig:intro_case}
% \caption{fig1}
\end{minipage}%
}%
\quad
\subfigure[]{
\begin{minipage}[!htb]{1\linewidth}
\centering % ,height=0.8\linewidth
\includegraphics[width=1\textwidth]{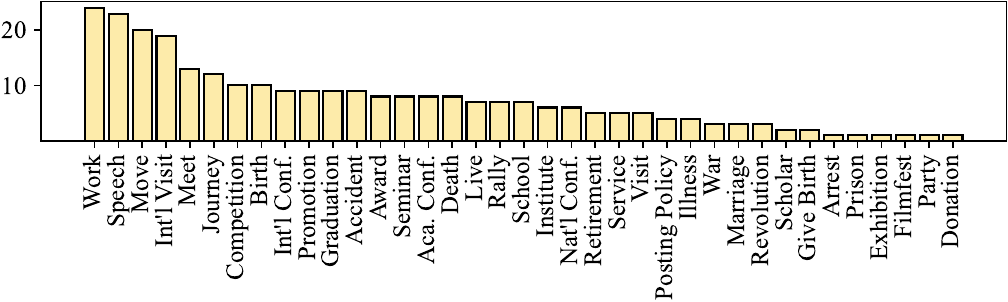}
%\caption{fig2}
\label{fig:intro_stat}
\end{minipage}%
}%
\centering
\caption{(a) Example of extracted triplets and their contexts, where green boxes represent correct trajectory information and red boxes represent incorrect one.
(b) The distribution of different trajectory types.
}
\label{fig: intro_fig}
\end{figure}

It is challenging to fully utilize Wikipedia for the trajectory extraction task, due to the variety and heterogeneity of the trajectories. If we directly apply Named Entity Recognition (NER) tools to detect elements and construct \textit{(person, time, location)} triplets, only 30\% of the triplets are entirely correct, according to our estimation. To improve this, context information should be considered. Snippets (1) and (2) in Figure~\ref{fig:intro_case} showcase the importance of context. Snippet (2) highlights an incorrect extraction candidate -- \textit{(Bob Hayes, 1964, USA)}, while if we have the sports event context, we would know the location should be Tokyo as shown in snippet (1), instead of the USA which \textit{Hayes} is representing. However, it is not easy to utilize the context information, since there is a large variety of contexts and different contexts may suggest different extraction patterns. We examine trajectories on 10 biography pages (out of the 1,930,519 ones in total) and the results already contain 37 categories. As shown in Figure~\ref{fig:intro_stat}, the contexts suggest a relatively even distribution, with various types covering 83\%, apart from work and speech. Therefore, for any work that uses the contexts for such triplet extractions, the labeled training data would inevitably contain a very small portion of all possible context types. This would be problematic when classifying triplets with unfamiliar contexts, giving rise to an overfitting/generalization issue. This may explain the low Recall of the traditional rule-based method for a similar task~\cite{lucchini2019following}.
In the study by~\citet{vempala2020extracting}, samples (also with contexts) were labeled from only 100 biography pages, focusing solely on the trajectories of the person who is the subject of the biography.
The F1 score was $74\%$ and if they test on samples from other people, the performance is expected to be lower.
The generalizability may be improved, if the classifier knows what the contexts of triplets in the wild look like beyond the very limited labeled triplets, which rhymes with the idea of semi-supervised learning~\cite{van2020survey}.

Despite the heterogeneity of the trajectories suggested in Figure~\ref{fig:intro_stat}, similar contexts often imply similar extraction rules and vice versa. For instance, the contexts of snippets (1) and (3) in Figure~\ref{fig:intro_case} are similar (both about sport events), suggesting the same way of extraction, while the dissimilarity between the context of snippet (1) and that of snippet (4) (about birth and study) indicates the way of extraction for snippet (1) does not apply to snippet (4). Therefore, capturing the similarity and dissimilarity between training samples may help improve the classifier and contrastive learning shares this ideology. It creates pairs of similar examples and dissimilar examples and trains a model to distinguish between them~\cite{khosla2020supervised}.

It is worth noting that besides Wikipedia, news articles are another popular source for similar extraction tasks~\cite{piskorski2020timelines,peng2024did}. However, apart from the apparent differences such as shorter duration (mostly after the 17th century) and somehow biased coverage towards certain people such as politicians, artists and athletes~\cite{gebhard2020polusa}, news articles have their distinct characteristics. For a particular trip of a celebrity, there may be several news articles covering it and the key elements may be scattered in one article and among articles~\cite{peng2024did}. For Wikipedia's biographies that we work on, one trip usually appears only on one page and its descriptions are often clustered in one place on that page. Therefore, a previous framework of extracting celebrity trips from news articles, CeleTrip~\cite{peng2024did}, relying on capturing long-range cross-document dependency between key elements within one article and across articles with Word-Article graph, may not work well in our scenario. Instead, local semantics may be effective here.

%%%%%%%%%%%%%%%%%%%%%%%%%%
%%%%%%%%%%%%%%%%%%%%%%%%%%

In this paper, we propose \textbf{COSMOS} (\textbf{CO}ntrastive learning and \textbf{S}emi-supervised learning \textbf{MO}del for extracting \textbf{S}patio-temporal life trajectory) to accomplish the task of trajectory extraction from Wikipedia.
Before it goes to work, we extract the \textit{(person, time, location)} triplets as candidates for it to classify. A correct triplet means that this \textit{person} is actually in this \textit{location} at this \textit{time}. As mentioned above, in a person's biography, there could be $m$ names, $n$ times, and $k$ locations and this gives us a maximum of $mnk$ combinations as candidate triplets. To reduce the problem space, we take the heuristic to restrict the elements of a triplet to be in the same sentence and apply a syntax tree to further remove unreasonable combinations, as suggested by~\citet{temporal-extraction-2020}. According to our estimation, the remaining triplets would cover most of the correct ones.
These triplets, combined with their corresponding contexts are then fed into \textbf{COSMOS}.
COSMOS will classify these candidate triples into two categories (`trajectory' or `not trajectory') based on the triples themselves and their context.
The representation of a triplet and its context is obtained from BERT~\cite{devlin2018bert}.
As discussed earlier, to mitigate overfitting and to exploit the dis/similarity among training samples, we combine semi-supervised learning~\cite{van2020survey} and contrastive learning~\cite{khosla2020supervised}, in a joint training framework.

Additionally, for this task, we create a first dataset annotating the general life trajectories, \textit{WikiLifeTrajectory}, containing 8,852 triplets.

We summarize our contributions as follows.

\begin{itemize}
\item We formally formulate the task for life trajectory extraction from Wikipedia, and construct a curated dataset \textit{WikiLifeTrajectory} dataset for this task. Though we perform our experiments on Wikipedia biographies, similar methods can be extended to other biography content.
\item We design an effective method, \textbf{COSMOS}, which combines the idea of contrastive learning and semi-supervised learning to improve the generalization of the model facing diverse trajectory contexts. \textbf{COSMOS} extracts life trajectories and outperforms all baselines with an F1 score of 85.95\% on the dataset.
\item We make our framework, the million-level extracted trajectories, and the \textit{WikiLifeTrajectory} dataset publicly available\footnote{\url{https://github.com/ZhangDataLab/COSMOS}. For now, we release trajectories extracted from 5\% biography pages. The remaining data and the \textit{WikiLifeTrajectory} dataset will be released with the camera-ready version.}. 
The dense and interactive trajectories with wide coverage will support mobility analysis from case studies to large-scale modeling.
Besides, we conduct an empirical analysis on the trajectories of historians to show the potential of our data. 
\end{itemize}

%%%%%%%%%%%%%%%%%%%%%%%%%%%

\section{Related Work}

\subsection{Life Trajectory Analysis}
Life trajectory data has extensive applications in the field of social sciences~\cite{yen2021personal}. For instance,~\citet{schich2014network} utilize the birth and death locations of more than 150,000 historically notable individuals to explore the evolution of culture centers.
In addition to cultural history, trajectory information of specific populations is also used for analysis for various purposes. The trips of U.S. presidents are analyzed to understand presidential policy~\cite{kernell2006going}, while the migration patterns of artists can serve government decision-making~\cite{Hautala2019artist}.
Life trajectories are also helpful in understanding human behavioral patterns. For example, ~\citet{Kleinepier2015LifePO} investigate the life trajectories of Polish immigrant families to identify factors influencing different family paths.

However, the current life trajectory data suffers from limitations in temporal scope, spatial coverage and inter-person interaction.
These limitations emphasize the monotonous nature of the available data, which lacks diversity and comprehensiveness, restricting the ability of existing life trajectory data for large-scale, fine-grained, and networked analysis.

\subsection{Spatio-Temporal Knowledge Extraction}

There have been several studies trying to extract spatial or temporal knowledge related to people from different information sources.
The most relevant to our task is the study working on text corpus. A relatively early study detects the type and timing of users' life events from social media based on hand-crafted features and traditional machine learning models~\cite{dickinson2015identifying}.
With the development of neural networks such as convolutional neural networks (CNNs) and recurrent neural networks (RNNs), researchers use deep learning models like Bi-LSTM to build life logs with timelines for individual users~\cite{yen2019personal}.

Recently, there are also some efforts dedicated to the extraction of spatio-tempoal knowledge.
\citet{lucchini2019following} extract births, deaths, and migrations of notable people on year-level.
Their approach focuses on extracting knowledge from semantic roles defined by FrameNet, and only considers 29 frames ``related to movements'', resulting in a low Recall in general life trajectories~\cite{lucchini2019following}.
\citet{peng2024did} detect celebrity trips from the news by modeling the news articles as graphs to capture the long-range cross-document dependency.
However, trajectories in Wikipedia are more compact within single biography pages, which may make the sequential models more suitable than graph-based models for modeling text.
In addition, \citet{vempala2020extracting} extract spatial timelines of the top 50 wealthiest individuals and top 50 multiple Olympic medalists from their biography pages based on LSTM. 
When applied to larger population groups and more diverse trajectory types, the model's generalization ability may be constrained if relying solely on limited labeled data.
To mitigate this limitation, we introduce additional unlabeled samples during model training and guide the model to classify the samples through comparison.

\section{Problem Statement}

Before we formalize the core problem, we describe the heuristic which is briefly mentioned in the Introduction.
Since we want to find where and when someone has appeared on a Wikipedia page, we first retrieve all possible combinations of \textit{(person, time, location)} from the page as candidate triplets and then identify triplets representing someone's trajectory.
To obtain these candidate triplets, we employ NER tools to recognize all entities related to person, time and location on one page, following by combining them to \textit{(person, time, location)} triplets based on the structures of sentences, accompanied by the paragraphs they come from.
The task now becomes: for each candidate triplet \textit{(person, time, location)}, decide whether \textit{person} has actually appeared in \textit{location} at \textit{time}, given the corresponding paragraph.

Given a triplet $t=(person, time, location)$ and its corresponding paragraph $p$, we train a model $f$ with trainable parameters $\Theta$ to classify $t$ into binary category $y \in \left\{0,1\right\}$, with $y=1$ indicating that $person$ has appeared in $location$ at $time$, and $y=0$ being the opposite:
\begin{align}
    f : \left\{t, p, \Theta\right\} \rightarrow y.
\end{align}

Our pipeline for extracting candidate triplets will be described in Section Dataset.

\section{Dataset}
To the best of our knowledge, there is no dataset annotating the general life trajectories we are interested in, so we create our own ground truth dataset. This section describes how we construct the \textit{WikiLifeTrajectory} dataset, which consists of two manually annotated datasets, namely ``Regular'' and ``Representative''.

\subsection{Data Collection}
The data is from the biography pages of the English Wikipedia. The list of people and the corresponding links to the Wiki pages are from Wikidata~\cite{moller2021survey}. In all, we have 1,930,519 people and their biography pages.

\subsection{Data Processing and Labeling}
This section describes how we extract and annotate candidate triplets in \textit{(person, time, location)} format from biography pages.

\subsubsection{Extracting Triplets}
We design a pipeline to extract candidate triplets from biography pages. The usability of this pipeline is also evaluated in real-world scenarios at the end of this section.
Previous research utilizes NER tools to identify entities in sentences and employs a sliding window approach to select time and location entities related to a person~\cite{vempala2020extracting}.
However, this method ignores the structure of sentences and will introduce many meaningless connections between entities~\cite{Liu2019ExtractingTP}.
To address this limitation, we draw inspiration from~\citet{temporal-extraction-2020} and construct parse trees to connect entities identified by NER.
Ultimately, we obtain candidate triplets in the form of \textit{(person, time, location)}.

In our extraction pipeline, we first use SpaCy\footnote{\url{https://spacy.io/}} to identify entities. We retain sentences that only contain time and location entities as target sentences to ensure that we can connect these entities by parse trees.
For each target sentence, we classify entities into four sets using SpaCy's classifications: \textit{person} (personal pronoun and name entity PERSON), \textit{time} (name entities DATE, TIME, and DURATION), \textit{location} (name entities GPE, LOC, EVENT, FAC, and ORG), and \textit{verb}.
In order to find the logical connection between entities, we refer to the previous approach~\cite{temporal-extraction-2020}. In each target sentence, we construct multiple (person, verb), (time, verb), and (location, verb) pairs based on the entities.
We define a parse-tree-based distance metric to estimate the relevance of different pairs. The distance is calculated as the minimum path from each entity in a pair to their lowest common ancestor (LCA), and the pair with the minimum distance is considered the most relevant.
Based on this comparison method, for each person entity in the \textit{person} set, we identify the relevant (person, verb*) pair. According to verb*, we determine the most relevant time and location entities. Finally, the verb entity serves as a bridge connecting these three entities, giving us candidate triplets of the form \textit{(person, time, location)}.

Finally, we compare the number of trajectories mentioned in the original pages with those extracted from the target sentences to measure the coverage of our method.
Four biography pages with a total of 106 trajectory descriptions are manually checked, and our extraction pipeline can cover at least 85\% (specifically, 86.11\%, 90.00\%, 85.71\%, and 86.96\% for each page, respectively) of the trajectories mentioned on different pages.
The uncovered trajectories contain ambiguous expressions of time or location, such as ``several years later'', which inherently represent vague trajectories and are even hard to recognize by humans.

\subsubsection{Annotating Triplets}
%%%%%%%%%%%%%%%%%%%%%%%%
\begin{figure}[!htb]
    \centering
    \includegraphics[width=1\linewidth]{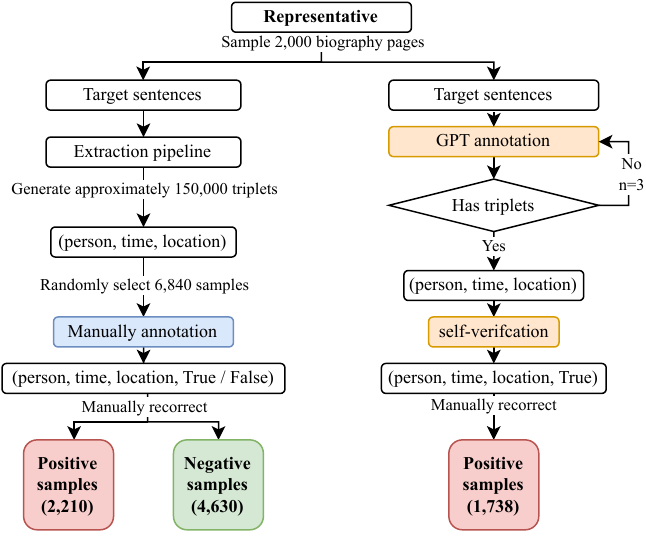}
    \caption{The flowchart illustrates the process of annotating the ``Representative'' dataset to obtain triplets and their corresponding labels.}
    \label{fig:annotation}
\end{figure}
%%%%%%%%%%%%%%%%%%%%%%%%%%%

As mentioned in the Introduction, we collect trajectories from 10 biography pages and find that trajectories vary due to the variety of occupations and life stages.
These trajectories are manually annotated in a \textit{(person, time, location)} triplet format and labeled as positive ($y=1$, defined in Problem Statement), resulting in the dataset ``Regular''.
In addition, to cover more representative trajectory types when constructing the ground truth, we sample and annotate the triplets extracted by our extraction tool to obtain the dataset ``Representative''.

Given that trajectory triplets' contexts are often associated with occupations~\cite{li2017spot}, we employ a stratified sampling and annotation strategy based on occupation to collect representative trajectories.
We first identify the top 300 occupations based on their occurrence frequencies from people we collect, and then select additional 2,000 biography pages through stratified sampling based on the proportion of each occupation.

After extracting the candidate triplets through our extraction tool, we adopt a mixed annotation approach utilizing both human annotators and GPT-3.5, as recent studies utilizing GPT for data annotation have piqued our curiosity regarding the effectiveness of GPT-based data labeling~\cite{thapa2023humans}.
From biography pages in ``Representative'', we choose individuals with longer page content and feed target sentences we extract to GPT-3.5 for annotation.
Drawing inspiration from previous studies~\cite{Wang2023GPTNERNE,han2023information}, we design an instructive prompt\footnote{See the Appendix for the prompts we provide for GPT.} for GPT to generate trajectory triplets from the given target sentences following a self-verification mechanism. Though more than half of the trajectories will be lost after GPT's self-verification, the Precision of resulting annotations exceeds 90\%.

Ultimately, we obtain 1,738 triplets labeled as positive ($y=1$) annotated by GPT. Additionally, three undergraduate students with English proficiency annotate a total of 6,840 triplets, using target sentences distinct from those fed to GPT, where 2,210 triplets are labeled as positive and 4,630 triplets are labeled as negative ($y=0$). A graduate student then verifies these results. The corresponding workflow is shown in Figure~\ref{fig:annotation}.

After the above annotation process, the ``Representative'' consists of 3,948 positive triplets and 4,630 negative triplets, while the ``Regular'' consists of 274 positive triplets. The triplets, along with the paragraphs and biography pages they originated from, are combined to form a complete JSON formatted dataset (i.e. the \textit{WikiLifeTrajectory} dataset). Additionally, we randomly extract 50,000 unlabeled triplets from biography pages for subsequent semi-supervised learning in our method.

\section{Method}

\begin{figure}[!htb]
    \centering
    \includegraphics[width=\linewidth]{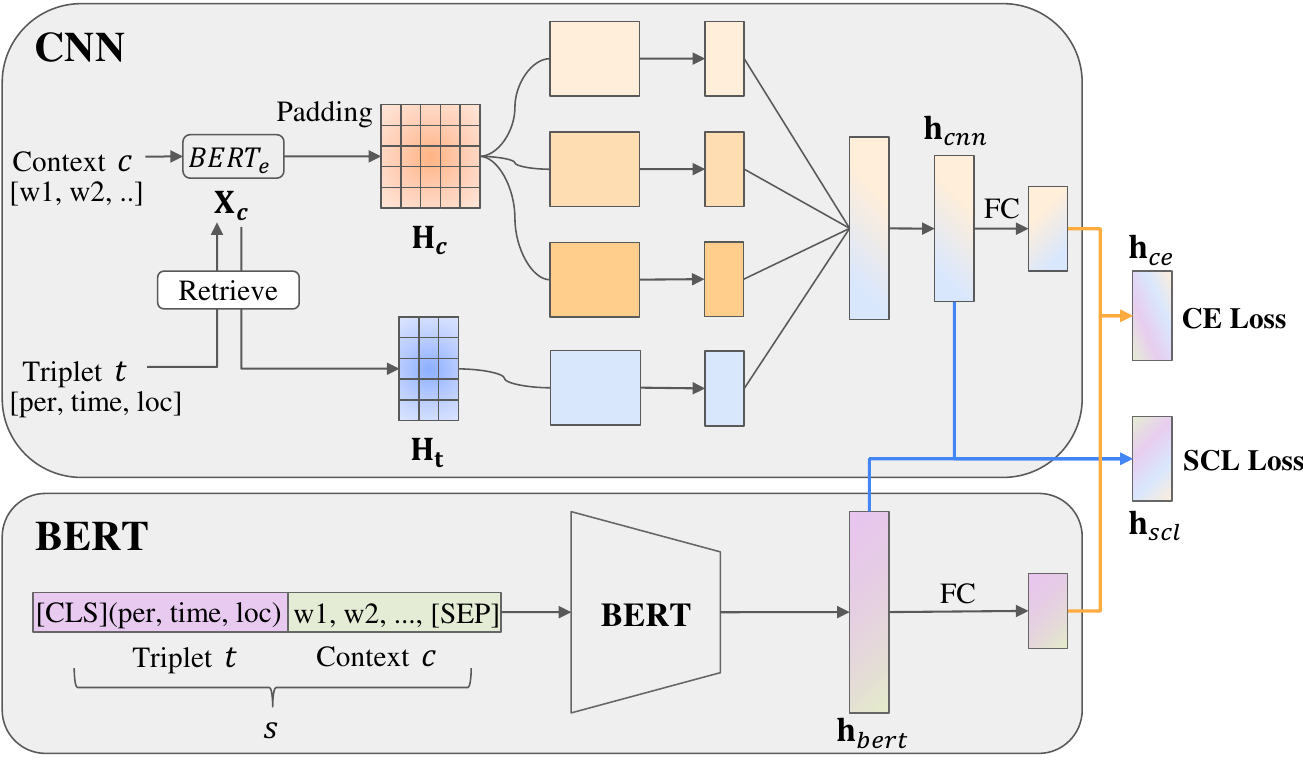}
    \caption{The framework of COSMOS. COSMOS learns the representations of triplets and their contexts through parallel CNN and BERT, and then classifies them based on the resulting representations.}
    \label{fig:model}
\end{figure}

The structure of COSMOS is shown in Figure~\ref{fig:model}.
As described in Introduction and Problem Statement, for any triplet $t=(person, time, location)$, COSMOS learns its overall representations ($\mathbf{h}_{ce}$) through an ensemble model and then predicts the triplet label.
When training the model, we introduce a combination of contrastive learning and semi-supervised learning.
The following sections elaborate (1) the design of the ensemble model and how we integrate (2) contrastive learning and (3) semi-supervised learning into our model.

%%%%%%%%%%%%%%%%%%%%%%%%
%%%%%%%%%%%%%%%%%%%%%%%%%%%

\subsection{Representation of Triplet}

In life trajectory extraction scenarios, achieving a balance of Recall and Precision is crucial for accurate and reliable extraction. However, during our experimental analysis, we observe that individual models based on CNN or BERT exhibit contrasting performance characteristics (see Experimental Results). While the CNN model demonstrates a higher Precision, resulting in more precise predictions, the BERT model exhibits a higher Recall, capturing more relevant information. This discrepancy in performance motivates us to explore the potential benefits of integrating these two models into a hybrid approach.

Given a triplet $t$ and its corresponding paragraph $p$, we will first match the sentences, which contain the target time or location in $t$, from $p$ to compose the triplet's context $c$.
Context $c$ is specifically represented as a sequence of words.
We then combine $t$ with its context $c$ using CNN and BERT in parallel (consisting of a joint training pipeline) to learn an overall representation of $t$.

\subsubsection{Representation from CNN}
We first use the pre-trained English BERT model (denoted as $BERT_e$ in Figure~\ref{fig:model}) to extract word embeddings of $c$ and obtain $\mathbf{X}_{c} \in \mathbb{R}^{n \times d}$, where $n$ is the sequence length of $c$ and $d$ is the dimensions of embeddings. Then, we pad $\mathbf{X}_{c}$ to $\mathbf{H}_{c} \in \mathbb{R}^{k_1 \times d}$.
Meanwhile, we retrieve the embeddings of the words corresponding to each element in triplet $t$ (i.e. person, time and location) from $\mathbf{X}_{c}$ to obtain $t$'s representation $\mathbf{H}_{t} \in \mathbb{R}^{3 \times d}$.

After that, $\mathbf{H}_{c}$ is independently passed through three parallel Conv2d layers, each with their respective learnable weights. Let $\mathbf{h}_{c}^{i}, i \in \{1, 2, 3\}$ represent the outputs from each Conv2d layer. On the other hand, $\mathbf{H}_{t}$ undergoes a single Conv2d operation to get $\mathbf{h}_{t}$.
Then we concatenate them to obtain $\hat{\mathbf{h}_{cnn}}$ (Eq.~\ref{eq: conv2d}-\ref{eq: concat conv}). Note that all the above Conv2d Layers have the same output dimension.

\begin{align}
    \mathbf{h}_{*} &= \textrm{Conv2d}(\mathbf{H}_{*}),\label{eq: conv2d} \\
    \hat{\mathbf{h}_{cnn}} &= \mathbf{h}_{c}^{1} \oplus \mathbf{h}_{c}^{2} \oplus \mathbf{h}_{c}^{3} \oplus \mathbf{h}_{t},\label{eq: concat conv}
\end{align}

Finally, we get the representation $\mathbf{h}_{cnn}$ from CNN through a linear layer (Eq.~\ref{eq: cnn output}).

\begin{align}
    \mathbf{h}_{cnn} &= \mathbf{w}_{c}^\top \hat{\mathbf{h}_{cnn}} + \mathbf{b}_{c},\label{eq: cnn output}
\end{align}
where $\mathbf{w}_{c}$ is the weight parameter.

\subsubsection{Representation from BERT}
We put $t$ at the beginning of $c$, forming a complete input sequence $s$. The resulting $s$ is then fed into the BERT model for fine-tuning, obtaining a combination representation $\mathbf{h}_{bert}$ (Eq.~\ref{eq: bert output}).

\begin{align}
    \mathbf{h}_{bert} &= \textrm{BERT}(s).\label{eq: bert output}
\end{align}

\subsubsection{Joint Training of CNN and BERT}
To jointly train CNN and BERT, we first transform $\mathbf{h}_{cnn}$ and $\mathbf{h}_{bert}$ through two independent learnable linear layers with the same output dimension $k_2$, resulting in $\mathbf{h}_{cnn}^{'}$ and $\mathbf{h}_{bert}^{'}$, respectively. Then, $\mathbf{h}_{cnn}^{'}$ and $\mathbf{h}_{bert}^{'}$ are concatenated into $\mathbf{h}_{ce}$ for classification (Eq.~\ref{eq: cnn trans1}-\ref{eq: softmax}).

\begin{align}
    \mathbf{h}_{cnn}^{'} &= \mathbf{w}_{cnn}^\top \mathbf{h}_{cnn} + \mathbf{b}_{cnn} \label{eq: cnn trans1}, \\
    \mathbf{h}_{bert}^{'} &= \mathbf{w}_{bert}^\top \mathbf{h}_{bert} + \mathbf{b}_{bert} \label{eq: bert trans1}, \\
    \mathbf{h}_{ce} &= \mathbf{h}_{cnn}^{'} \oplus \mathbf{h}_{bert}^{'}, \label{eq: ce output} \\
    y_{i}^{'} &= \textrm{Softmax}(\mathbf{h}_{ce}), \label{eq: softmax}
\end{align}
where $\mathbf{w}_{*}$ is the weight parameter.

Additionally, we also preserve a combination representation for calculating the contrastive learning term (see the next section). Specifically, $\mathbf{h}_{cnn}$ and $\mathbf{h}_{bert}$ are individually passed through another two learnable linear layer with the same output dimension $k_3$, resulting in $\mathbf{h}_{cnn}^{\diamond}$ and $\mathbf{h}_{bert}^{\diamond}$, respectively. These representations are then concatenated and passed through an attention layer to obtain a combined feature vector $\mathbf{h}_{scl}$ (Eq.~\ref{eq: scl output}).

\begin{align}
    \mathbf{h}_{scl} &= \textrm{attn}( \mathbf{h}_{cnn}^{\diamond} \oplus \mathbf{h}_{bert}^{\diamond}), \label{eq: scl output}
\end{align}
where attn is the attention layer.

\subsection{Supervised Contrastive Learning Loss}

Traditional classification tasks usually train classification models only with cross-entropy loss~\cite{khosla2020supervised}. However, cross-entropy loss can lead to instability and poor generalization when labeled data is limited~\cite{zhang2018generalized}. Supervised contrastive learning is proposed to combine the idea of ``learn to compare'' from contrastive learning into the supervised setting, alleviating the shortcomings of cross-entropy loss overly relying on the label itself.~\cite{khosla2020supervised}.

With the output feature vector $\mathbf{h}_{scl}$, we next compute the supervised contrastive learning loss term in Eq.~\ref{eq: scl loss}, following~\citet{gunel2020supervised}.

\begin{align}
    \mathcal{L}_{SCL} &= \sum_{i=1}^{N} 
    \frac{-1}{N_{y_i}-1}
    \sum_{j=1}^{N}\mathbf{1}_{i\neq j}\mathbf{1}_{y_i \neq y_j}
    \log{\frac{e_{ij}}{\sum_{k=1}^{N}\mathbf{1}_{i \neq k}e_{ik}}}, \notag \\
    e_{mn} &= \exp(h_{scl}^{m} \cdot h_{scl}^{n}/\tau), \label{eq: scl loss}
\end{align}

where $N$ is the batch size, $h_{scl}$ is the feature vector from the ensemble model, $N_{y_i}$ is the total number of examples in the batch that have the same label as $y_i$, and $\tau$ is an adjustable temperature parameter. We combine cross-entropy loss with $\mathcal{L}_{SCL}$ to calculate the loss $\mathcal{L}_{S}$ on labeled data with a scalar weighting hyperparameter $\lambda$ (Eq.~\ref{eq: supervised loss}-\ref{eq: ce loss}).

\begin{align}
    \mathcal{L}_{S} &= (1-\lambda)\mathcal{L}_{CE} + \lambda\mathcal{L}_{SCL}, \label{eq: supervised loss}\\
    \mathcal{L}_{CE} &= -\frac{1}{N} \sum_{i}^{N}(y_{i}  \log(y^{'}_{i})+(1-y_{i}) \log(1-y^{'}_{i})). \label{eq: ce loss}
\end{align}

\subsection{Semi-supervised Learning Loss}

Semi-supervised learning is a learning paradigm that utilizes unlabeled data to enhance the performance of the model, particularly in scenarios where there is limited labeled data but a large amount of unlabeled data~\cite{van2020survey}. In our scenario, labeling triplets is time-consuming and difficult to ensure a sufficient number of long-tail trajectory contexts, while the triplet extraction pipeline can quickly generate large amounts of unlabeled data. Therefore, we introduce semi-supervised learning during training to enhance the generalization ability of the model through unlabeled data.

Pseudo-label~\cite{lee2013pseudo} is a simple but efficient method performing semi-supervised learning for deep neural networks. The principle behind Pseudo-label is to take the model's prediction on unlabeled data as pseudo-labels, and compute the corresponding loss of unlabeled data. This loss is then combined with the loss computed on labeled data. We adopt the idea of Pseudo-label to shift our model from supervised to semi-supervised setting by computing $\mathcal{L}_{U}$. Here we use cross-entropy loss instead of the combination of $\mathcal{L}_{SCL}$ for unlabeled data, since the predicted label may not ensure that both positive and negative labels appear in each batch, which is the prerequisite for calculating $\mathcal{L}_{SCL}$.

\begin{align}
    \mathcal{L} &= \mathcal{L}_{S} + \alpha(b, t) \mathcal{L}_{U}, \label{eq: final loss} \\
    \mathcal{L}_{U} &= \mathcal{L}_{CE}, \\
    \alpha(b, t) &=
        \begin{cases}
            0 & b \leq c_{1}B \\
            \frac{b}{B} \cdot \frac{\gamma}{t+1} & c_{1}B < b \leq c_{2}B \\
            1 & b > c_{2}B \\
        \end{cases}.
\end{align}

where $b$ and $t$ are ordinal numbers meaning the $b$-th batch in the $t$-th epoch, $B$ is the number of total batches in each epoch, $c_1$, $c_2$, and $\gamma$ are hyperparameters. $\alpha(b, t)$ is used to balance $\mathcal{L}_{U}$ and $\mathcal{L}_{S}$.

\section{Experiments}

\subsection{Train/Test Split} We divide the ``Representative'' dataset into training and testing in a 7:3 ratio. Given the relatively independent nature of the annotated triplets obtained through sampling, a random split is less likely to result in data leakage. Additionally, we randomly sampled 20\% from the training set as a validation set for hyperparameter tuning. Meanwhile, we preserve the ``Regular'' dataset as an independent test set.

\subsection{Implementation Details}
In this paper, we use a BERT-base\footnote{\url{https://github.com/google-research/bert}} model to generate word embeddings, while any suitable model can be employed as well. In COSMOS, we set $d=768$, $k_{1}=100$, $k_{2}=2$, and $k_{3}=32$. We set $\lambda=0.2$, $\tau=0.1$, $c_1=0.1$, $c_2=0.9$, and $\gamma=0.8$ in loss terms. The hyperparameters including $\lambda$, $\tau$, and $\gamma$ are adjusted with grid search~\cite{bergstra2011algorithms}.

To train COSMOS, we set the learning rate to $5e^{-5}$ and use the Adam optimizer~\cite{kingma2014adam}. Moreover, we adopt an early stop strategy to avoid overfitting. We conduct our experiments with two RTX 3090.

\subsection{Evaluation Metrics}

To quantitatively evaluate our model, we use the following performance metrics.

\subsubsection{Metrics for Prediction Performance}
We use Accuracy (Acc), Precision (P), Recall (R), and F1 score to evaluate the model's ability to extract trajectory triplets by test set from ``Representative''. The F1 score holds significant importance in our scenario as it represents the balance of Precision and Recall, which has a great impact on downstream analysis.
Given the mixed origins of our dataset samples (manually labeled and GPT-labeled), we denote the respective sub dataset with subscripts for distinction. ``Representative'' refers to all the samples from both sources, ``Representative$_\textrm{m}$'' denotes manually labeled samples, and ``Representative$_\textrm{g}$'' represents GPT-labeled samples. Given that GPT is only used for labeling positive samples, we only calculate the Recall for GPT-labeled samples.

\subsubsection{Metrics for Coverage Performance}
We use Recall (R) to assess the coverage performance of the model on all the trajectories within each biography page by ``Regular''. Additionally, we compute the average Recall and its standard deviation across different pages.

\subsection{Baseline Methods} Here we introduce seven baselines. The first four models, LR (TF-IDF), CNN, Bi-LSTM, and CeleTrip, as mentioned in Related Work, are used in the previous work to extract spatio-temporal knowledge about people from text corpus and related extraction tasks.
The next three, BERT, RoBERTa, and GPT-3.5, are language models commonly used for general language tasks. Recently, GPT-3.5 has gained significant attention as a general-purpose language model and has been evaluated in various information extraction tasks~\cite{han2023information,gao2023exploring}.
However, these studies have indicated that there is still a gap between the performance of GPT and the supervised SOTA in information extraction tasks such as Event Detection.

Since we use both the triplet and its context in COSMOS, as a fair comparison, we combine the triplet and its context as an entire input to each baseline.

\begin{itemize}
    \item \textbf{LR (TFIDF)~\cite{dickinson2015identifying}:} The researchers use TFIDF vectors for text representation on their classification task. We use a similar approach to extract representations and employ Logistic Regression as the classifier in our experiments.

    \item \textbf{CNN~\cite{nguyen2017robust}:} During their research into identifying crisis events from personal tweets, the model captures the local semantic features of text. The resulting features are then combined together and utilized as the input for the classifier. In our experimental setup, we employ the softmax classifier for all deep learning models.

    \item \textbf{Bi-LSTM~\cite{yen2019personal}:} When detecting real-life events from users' tweets, the researchers use Bi-LSTM to capture the sequential semantics within tweets content. In our task, we employ Bi-LSTM to model the triplets and their contexts.

    \item \textbf{CeleTrip~\cite{peng2024did}:} The researchers propose CeleTrip for detecting celebrity itineraries from news articles. CeleTrip models the location context as a word graph and utilizes Oriented Pooling to encode information of the target celebrity and location. Similarly, we model triplet's context as a graph and fuse triplet features through Oriented Pooling in our experiments.
    
    \item \textbf{BERT~\cite{devlin2018bert}:} We consider BERT as part of our baselines. BERT is a widely used pre-trained language model and has demonstrated strong performance across various language tasks. When fine-tuning BERT for classification, we follow the same hyperparameter suggested by~\citet{devlin2018bert}.

    \item \textbf{RoBERTa~\cite{liu2019roberta}:} As an enhancement of BERT, RoBERTa is trained with longer data and more data while removing the Next Sentence Prediction (NSP) objective. It has been shown to outperform BERT on many downstream tasks~\cite{liu2019roberta}. In the experiment, we use the same settings as BERT to fine-tune it.
    
    \item \textbf{GPT-3.5\footnote{\url{https://platform.openai.com/docs/models/gpt-3-5-turbo}}:} We introduce GPT-3.5 (gpt-3.5-turbo-0613) as another baseline for our task. See the Appendix for the prompt we provide for GPT.
    
\end{itemize}

Additionally, we conduct partial experiments on GPT-4\footnote{\url{https://platform.openai.com/docs/models/gpt-4-turbo-and-gpt-4}} (gpt-4-0613) due to the cost of API call, and the results and implementation details are reported in the Appendix.

\subsection{Experimental Results}

We evaluate the performance of models from two aspects. The ``Representative'' dataset is used to assess the prediction performance of models on sampled trajectory triplets, while the ``Regular'' dataset evaluates the model's coverage performance on life trajectories within complete biography pages.

\subsubsection{Prediction Performance}
The results of each model on datasets ``Representative'' and ``Regular'' are shown in Table~\ref{tab: experimental results}. Among these models, COSMOS achieves the best overall performance (F1=85.95\%), followed by RoBERTa.

%%%%%%%%%%%%%%%%%%%%%%%%%%%%%%%%%%%%%%%%%%%%%%%%%%%%%%%%
%%%%%%%%%%%%%%%%%% Experimental Results   %%%%%%%%%%%%%%
\begin{table*}[!htb]
\setlength\tabcolsep{3pt}
  \centering
    \begin{tabular}{cccccccccccc}
    \toprule
    % \midrule
    \multirow{3}{*}{} & \multicolumn{4}{c}{\textbf{Representative}} & \multicolumn{4}{c}{\textbf{Representative$_\textrm{m}$}} & \multicolumn{1}{c}{\textbf{Representative$_\textrm{g}$}} & \multicolumn{2}{c}{\textbf{Regular}} \\
    \cmidrule(lr){2-5} \cmidrule(lr){6-9} \cmidrule(lr){10-10} \cmidrule(lr){11-12}
    & Acc (\%) & P (\%) & R (\%) & F1 (\%) & Acc (\%) & P (\%) & R (\%) & F1 (\%) & R (\%) & R (\%) & Avg-R (std)\\
    \midrule
    GPT-3.5    & 63.99          & 56.53          & 95.12* & 70.91          & 55.00          & 41.48          & 91.39* & 57.06          & 100.00* & 92.33* & 0.9126 ± 0.0716\\
    LR (TFIDF)     & 74.47          & 75.45          & 66.24 & 70.55          & 75.67          & 62.62          & 63.64 & 63.13          & 69.64  & 44.52 & 0.4262 ± 0.1751\\
    CeleTrip   & 82.55          & 81.77          & 80.05 & 80.90          & 81.31          & 70.26          & 74.33 & 72.24          & 87.54  & 60.94 & 0.5614 ± 0.2351\\
    Bi-LSTM     & 84.38          & 81.38          & 85.77 & 83.52          & 81.94          & 69.66          & 79.37 & 74.20          & 94.16  & 75.18 & 0.7549 ± 0.2031\\
    CNN        & 84.42          & \textbf{84.91}          & 80.55 & 82.67          & 82.62          & \textbf{74.08}          & 72.10 & 73.08          & 91.63  & 63.50 & 0.6344 ± 0.2111\\
    BERT       & 84.65          & 80.10          & \textbf{88.80} & 84.23          & 82.08          & 68.39          & \textbf{84.12} & 75.44          & 94.94  & \underline{81.02} & 0.8304 ± 0.1398\\
    RoBERTa       & \underline{86.09}          & 82.88          & \underline{88.04} & \underline{85.38}          & \underline{83.68}         & 71.94          & \underline{82.19} & \underline{76.73}          & \textbf{95.71}  & 77.00 & 0.7389 ± 0.1583\\
    \textbf{COSMOS} & \textbf{86.79} & \underline{84.41} & 87.54 & \textbf{85.95} & \textbf{84.61} & \textbf{74.08} & 81.45 & \textbf{77.59} & \underline{95.52}  & \textbf{82.11} & 0.8169 ± 0.0906\\
\bottomrule
    \end{tabular}
    \caption{Performance comparison on the test set. Due to the extreme imbalance between Precision and Recall of GPT-3.5, we specifically highlight the Recall for it with an asterisk (*). Apart from that, the best results are indicated by bold text, while the second-best ones are highlighted with underlines.}
    \label{tab: experimental results}%
\end{table*}%
%%%%%%%%%%%%%%%%%%%%%%%%%%%%%%%%%%%%%%%%%%%%%%%%%%%%%%%%

As the model with the best overall performance, COSMOS integrates CNN and BERT, approaching CNN in Precision and approaching BERT in Recall, indicating its effective fusion of the advantages of both models.
The CNN model, which focuses on local semantics, achieves the best Precision (84.91\%), while the graph-based model CeleTrip, which captures long-distance semantics, has a lower performance (81.77\%).
The possible reason may be that in our scenario, the text is relatively short, and the semantics are more concentrated, making sequential models more suitable for text modeling than graph-based models.

Additionally, we notice that although the GPT-3.5 model achieves the highest Recall (95.12\%), it exhibits a severe imbalance between Precision (56.53\%) and Recall, introducing a significant amount of noise while retrieving life trajectories.
Given its tendency to misinterpret temporal and spatial information in the text as trajectories, it is not yet suitable for direct application in extracting life trajectories.
Meanwhile, BERT, which focuses on contextual understanding, achieves the highest Recall (88.80\%) among models other than GPT-3.5.
RoBERTa has a somehow unexpected lower Recall compared to BERT (decreased by 0.76\%), while overall outperforming BERT in our task (increased by 1.15\% in F1). Different from BERT, RoBERTa's design removes the NSP Loss during the training process, which might have an impact on downstream tasks that require language inference~\cite{devlin2018bert}.
The Bi-LSTM model, which also captures bidirectional semantic information like BERT and RoBERTa, achieves relatively higher Recall than other models.
This indicates that the capturing of continuous semantics contributes to retrieving more trajectories in our scenario.

The performance of these models on ``Representative$_\textrm{m}$'' is similar to those on ``Representative''. However, we observe an intriguing phenomenon in our experimental results, where all models except LR (TFIDF) exhibit a superior increase on Recall when tested on the samples labeled by GPT-3.5 (``Representative$_\textrm{g}$'').
To explore the possible reason for that, we compare these two datasets (``Representative$_\textrm{m}$'' and ``Representative$_\textrm{g}$'') by computing the distribution of verbs from the positive triplets' contexts.
The top 5 most frequent words (``born'', ``be'', ``die'', ``move'' and ``marry'') appear in the exact same order in both datasets, and accounts for 40.20\% and 44.55\%, respectively.
Then, we present the distribution of verbs from 6th to 10th in Table~\ref{tab: verb on testset}.
It seems that manually annotated data involves more personal life and behavior, while GPT-3.5 annotated data focuses more on work, responsibilities and performance.
This raises the question of why models using pre-trained word embeddings display consistently better learning outcomes on the GPT-3.5 annotated data, despite there being no significant difficulty difference in verbs between the two test sets.
Further investigation into the underlying factors influencing this phenomenon and its generalizability across different tasks and datasets presents an interesting direction for future research.

%%%%%%%%%%%%%%%%%%%%%%
\begin{table}[!htb]
% \small
  \centering
    \begin{tabular}{ll}\toprule
     \textbf{Representative$_\textrm{m}$} & \textbf{Representative$_\textrm{g}$} \\\midrule
    graduate (2.5\%) & perform (2.3\%)\\
    become (2.2\%) & make (2.3\%)\\
    live (2.0\%) & hold (1.5\%)\\
    take (1.6\%) & work (1.5\%)\\
    leave (1.6\%) & serve (1.5\%)\\
\bottomrule
    \end{tabular}%
    \caption{Top 6-10 high-frequency verbs in triplets' contexts.}
    \label{tab: verb on testset}%
\end{table}%
%%%%%%%%%%%%%%%%%%%%%%

\subsubsection{Coverage Performance}
We evaluate the models' coverage of life trajectories on individual Wikipedia biography pages using the ``Regular'' dataset. Most models have lower Recall on ``Regular'' dataset compared to the Recall on manually labeled samples in ``Representative'' dataset. 

Apart from GPT-3.5, only COSMOS shows a slight improvement (increased by 0.66\%) in this metric.
This indicates that the combination of contrastive learning and semi-supervised learning enhances the model's generalization ability. Additionally, excluding GPT-3.5, which sacrifices Precision for Recall, COSMOS achieves the highest Recall (82.11\%) and exhibits the least standard deviation across different pages. This suggests that COSMOS is practical in real-world applications.

\subsubsection{Examples of Trajectories Extracted}

\begin{figure}[!htb]
    \centering
    \includegraphics[width=\linewidth]{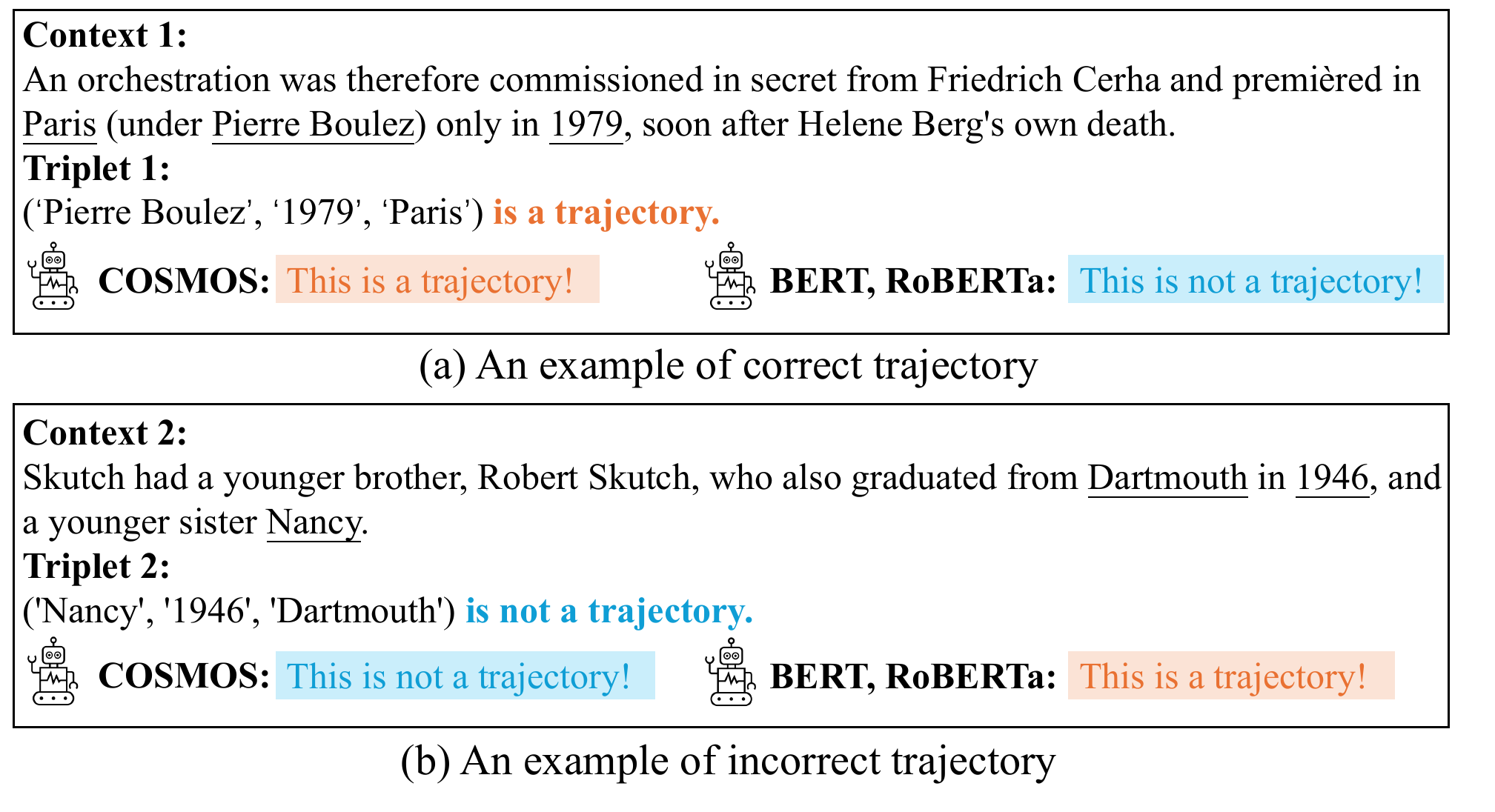}
    \caption{Examples of extracted trajectories. Underlines represent the corresponding element in the given triplet.}
    \label{fig:case_study}
\end{figure}

Alongside the quantitative analysis, we also examine the trajectories extracted by different models. 
To illustrate COSMOS's capacity for generalization across diverse trajectory descriptions and its precision in extracting information, we provide two representative examples below.
Figure~\ref{fig:case_study}~(a) is the trajectory of Pierre Boulez conducting premiere in Paris, described as \textit{``an orchestration ... (under Pierre Boulez)''}. COSMOS gives the correct classification, while BERT and RoBERTa miss this relatively implicit description of trajectory.
Meanwhile, Figure~\ref{fig:case_study}~(b) is an example of an incorrect trajectory, and the given triplet is \textit{(``Nancy'', ``1946'', ``Dartmouth'')}.
COSMOS identifies the triplet as not being a trajectory of \textit{``Nancy''} while the other two models seem to be misled by other people's trajectory.

%%%%%%%%%%%%%%%%%%%%%%%%
%%%%%%%%%%%%%%%%%%%%%%%%%%%

\subsubsection{Error Analysis} We perform an error analysis on COSMOS to analyze its limitations. Overall, there are two types of error: false positives (triplets that are mistaken for trajectory, related to Precision) and false negatives (trajectory triplets that are missed, related to Recall). We sample 20\% from each type, resulting in 34 false positives and 35 false negatives.

The main error reasons for false positives are (1) the lack of background knowledge (59\%) and (2) mismatched time (18\%). 
We demonstrate reason (1) by the following example from one biography page: \textit{``Viceroy of India, Lord Curzon, partitioned Bengal.''} The verb \textit{``partition''} in this context means the allocation of resources rather than implying that \textit{``Lord Cornwallis''} himself was present in \textit{``Bengal''}; this may require a well grasp of geopolitical knowledge to help the model understand its underlying meaning.
For reason (2), time expressions included in parentheses such as \textit{``1887''} from \textit{``the Artist's Wife (I Havedøren, 1887)''} can mislead the model in finding the actual date of trajectory.

False negatives are mainly caused by: (1) vague time explanation (29\%); and (2) diversity in writing style (26\%).
Time expressions such as \textit{``monthly''} and \textit{``the next few years''} sometimes distract the model from finding the trajectories.
On the other hand, complex sentences written by diverse editors~\cite{ren2017crowd} such as \textit{``Exactly the same injury, 44 years later, in \textbf{August 1986}, in \textbf{Afghanistan}, his grandson, a military intelligence sergeant, \textbf{Iliyas Daudi}, who was blown up by an Italian anti-personnel mine.''} are challenging. The writing style with numerous comma-separated phrases makes it difficult for the model to connect essential information and classify it as a positive trajectory.

\subsection{Ablation Study}

We remove certain loss terms from COSMOS to validate their effectiveness, and the results are shown in Table~\ref{tab: ablation results}. COSMOS$_\textrm{\ w/o\ ssl}$ removes the semi-supervised learning loss. Similarly, COSMOS$_\textrm{\ w/o\ scl}$ removes the supervised contrastive learning loss. COSMOS$_\textrm{\ w/o\ ssl\&scl}$ removes both the semi-supervised learning and supervised contrastive learning loss terms.

%%%%%%%%%%%%%%%%%%%%%%%%%%%%%%%%%%%%%%%%%%%%%%%%%%%%%%%%
%%%%%%%%%%%%%%%%%% Ablation Results   %%%%%%%%%%%%%%
\begin{table*}[!htb]
\setlength\tabcolsep{2.3pt}
  \centering
    \begin{tabular}{cccccccccccc}
    \toprule
    % \midrule
    \multirow{3}{*}{} & \multicolumn{4}{c}{\textbf{Representative}} & \multicolumn{4}{c}{\textbf{Representative$_\textrm{m}$}} & \multicolumn{1}{c}{\textbf{Representative$_\textrm{g}$}} & \multicolumn{2}{c}{\textbf{Regular}} \\
    % subtitle
    \cmidrule(lr){2-5} \cmidrule(lr){6-9} \cmidrule(lr){10-10} \cmidrule(lr){11-12}
    & Acc (\%) & P (\%) & R (\%) & F1 (\%) & Acc (\%) & P (\%) & R (\%) & F1 (\%) & R (\%) & R (\%) & Avg-R (std)\\
    \midrule
    COSMOS$_\textrm{\ w/o\ ssl\&scl}$  & 85.23 & 83.00 & \underline{85.52} & 84.24 & 82.66 & 71.62 & \underline{77.89} & 74.62 & \textbf{95.52} & 68.97 & 0.6955 ± 0.1791 \\
    COSMOS$_\textrm{\ w/o\ ssl}$ & 85.85 & \underline{85.64} & 83.33 & 84.47 & 83.83 & \underline{75.33} & 75.22 & 75.27 & 93.96 & 69.34 & 0.6636 ± 0.2479 \\
    COSMOS$_\textrm{\ w/o\ scl}$ & \underline{86.63} & \textbf{87.47} & 82.91 & \underline{85.13} & \textbf{84.80} & \textbf{78.07} & 74.48 & \underline{76.23} & 93.96 & \underline{71.89} & 0.6777 ± 0.2109 \\
    \textbf{COSMOS} & \textbf{86.79} & 84.41 & \textbf{87.54} & \textbf{85.95} & \underline{84.61} & 74.08 & \textbf{81.45} & \textbf{77.59} & \textbf{95.52} & \textbf{82.11} & 0.8169 ± 0.0906\\
\bottomrule
    \end{tabular}
    \caption{Results of the ablation study. Bold text indicates the best results, while underlined text represents the second-best ones.}
    \label{tab: ablation results}%
\end{table*}%
%%%%%%%%%%%%%%%%%%%%%%%%%%%%%%%%%%%%%%%%%%%%%%%%%%%%%%%%

As shown in Table~\ref{tab: ablation results}, COSMOS demonstrates the best overall performance as expected, while COSMOS$_\textrm{\ w/o\ ssl\&scl}$ performs the worst, indicating the effectiveness of the corresponding design. Compared to COSMOS$_\textrm{\ w/o\ ssl\&scl}$, the inclusion of contrastive learning (COSMOS$_\textrm{\ w/o\ ssl}$) or semi-supervised learning (COSMOS$_\textrm{\ w/o\ scl}$) alone leads to an increase in Precision (increased by 2.64\% and 4.47\%, respectively) and a decrease in Recall (decreased by 2.19\% and 2.61\%, respectively), but improving the overall performance of the model. However, when both are incorporated into the model, Precision and Recall achieve a better balance, and result in a significant improvement in coverage performance (increased by 13.14\% in Recall on ``Regular''). It is possible that the abilities of contrastive learning and semi-supervised learning are complementary. Contrastive learning enhances the discriminative ability of the model by learning the similarity between samples~\cite{khosla2020supervised}, while semi-supervised learning provides additional information by utilizing unlabeled data~\cite{van2020survey}. When they are combined, contrastive learning facilitates the selection of more discriminative features, while semi-supervised learning can expand the training set and increase data diversity, which may contribute to enhancing the model's performance.

\section{Analysis of a Sample Set}

Trajectories of antiquarians, scientists, artists, and historians have been a focus on cultural history studies~\cite{schich2014network,fu2014cultural,kaiser2018artist,long2018comparison}. In this section, we extract and analyze the trajectories of historians, to further validate our method and provide a taste of our dataset.

\subsection{Extracting and Processing Trajectories}

We use the set of historians identified by~\citet{laouenan2022cross} and retrieve their biography pages from Wikipedia. After extracting candidate triplets and classifying them by COSMOS, 
we clean and augment the extracted trajectories by utilizing Hugging Face's co-reference resolution tool\footnote{\url{https://github.com/huggingface/neuralcoref}} for disambiguating the names.
Additionally, we standardize time information and geocode location information using Nominatim\footnote{\url{https://nominatim.org/}}.

\subsection{Data Quality, Density and Variety}
\begin{figure}[!h]
    \centering
    \includegraphics[width=\linewidth]{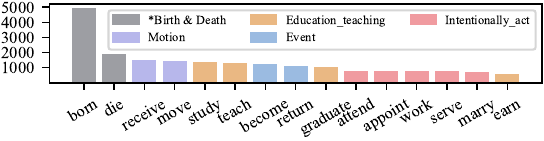}
    \caption{The distribution of the top 15 frequent verbs associated with the trajectories of historians.
    The horizontal axis represents verbs and the vertical axis represents their corresponding quantities. 
    The * legend indicates the custom category independent of FrameNet.
    }
    \label{fig:verb}
\end{figure}

\begin{figure}[!h]
    \centering
    \includegraphics[width=\linewidth]{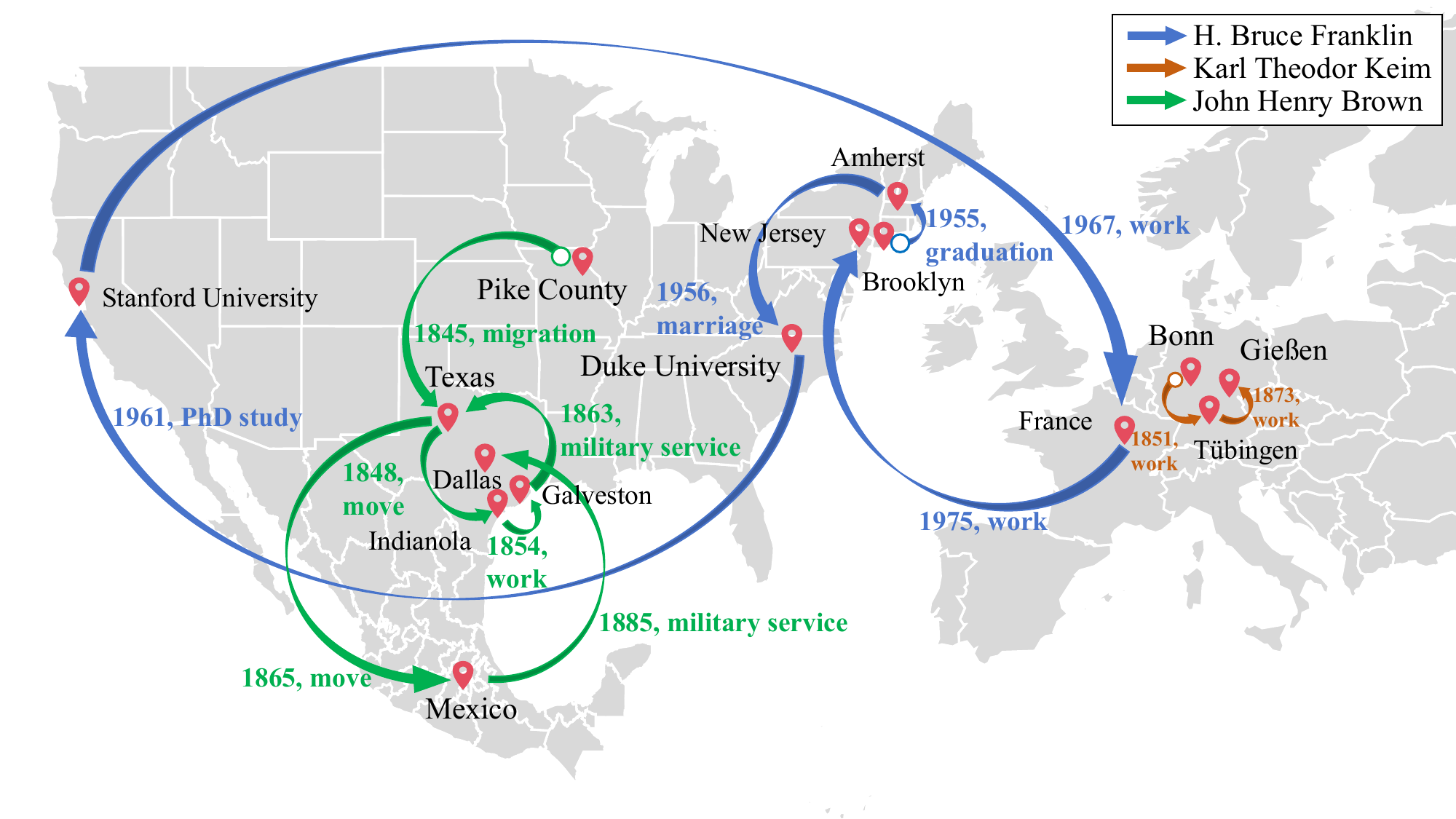}
    \caption{Life trajectories of \textit{H. Bruce Franklin}, \textit{Karl Theodor Keim} and \textit{John Henry Brown}. The arrows of each color represent the life trajectory of the corresponding individual. The start point of each trajectory is marked with a circle. The year and purpose of the move are labeled on the arrows.}
    \label{fig:map_case}
\end{figure}

After data processing, we obtain 20,786 trajectory triplets of 8,272 historians. 
We manually check 225 of the trajectory triplets and find that 80\% of them (180) are accurate, which is comparable with the model performance on the test set.
On average, each person has 2.51 triplets.
We further examine the types of trajectory triplets in this sample set. Similar to~\citet{lucchini2019following}, we use the verbs near the trajectory triplets to roughly represent the types according to the classification in FrameNet\footnote{\url{https://framenet.icsi.berkeley.edu/}} and Figure~\ref{fig:verb} shows the distributions of the top 15 verbs. Other than births and deaths, activities such as education, work, moving and travel take up 81.25\%, indicating that various interim points can add many new dimensions to datasets with only births and deaths.
Figure~\ref{fig:map_case} showcases the trajectories of three historians, \textit{H. Bruce Franklin} (5 triplets, in the 20th century), \textit{John Henry Brown} (6 triplets, in the 19th century), and \textit{Karl Theodor Keim} (2 triplets, in the 19th century). We can observe different movement patterns -- \textit{Keim}'s activities seem to have a limited geographic span, while \textit{Franklin} is an active traveler across continents, which may have benefited from the convenience of long-distance travel in the second half of the 20th century. \textit{Brown} had six moves within the U.S. and Mexico between 1845 and 1885, spending most of his life in Texas. The movements can surely be further quantified and aggregated over more historians to seek meaningful insights. 

\begin{figure}[!h]
    \centering
    \includegraphics[width=\linewidth]{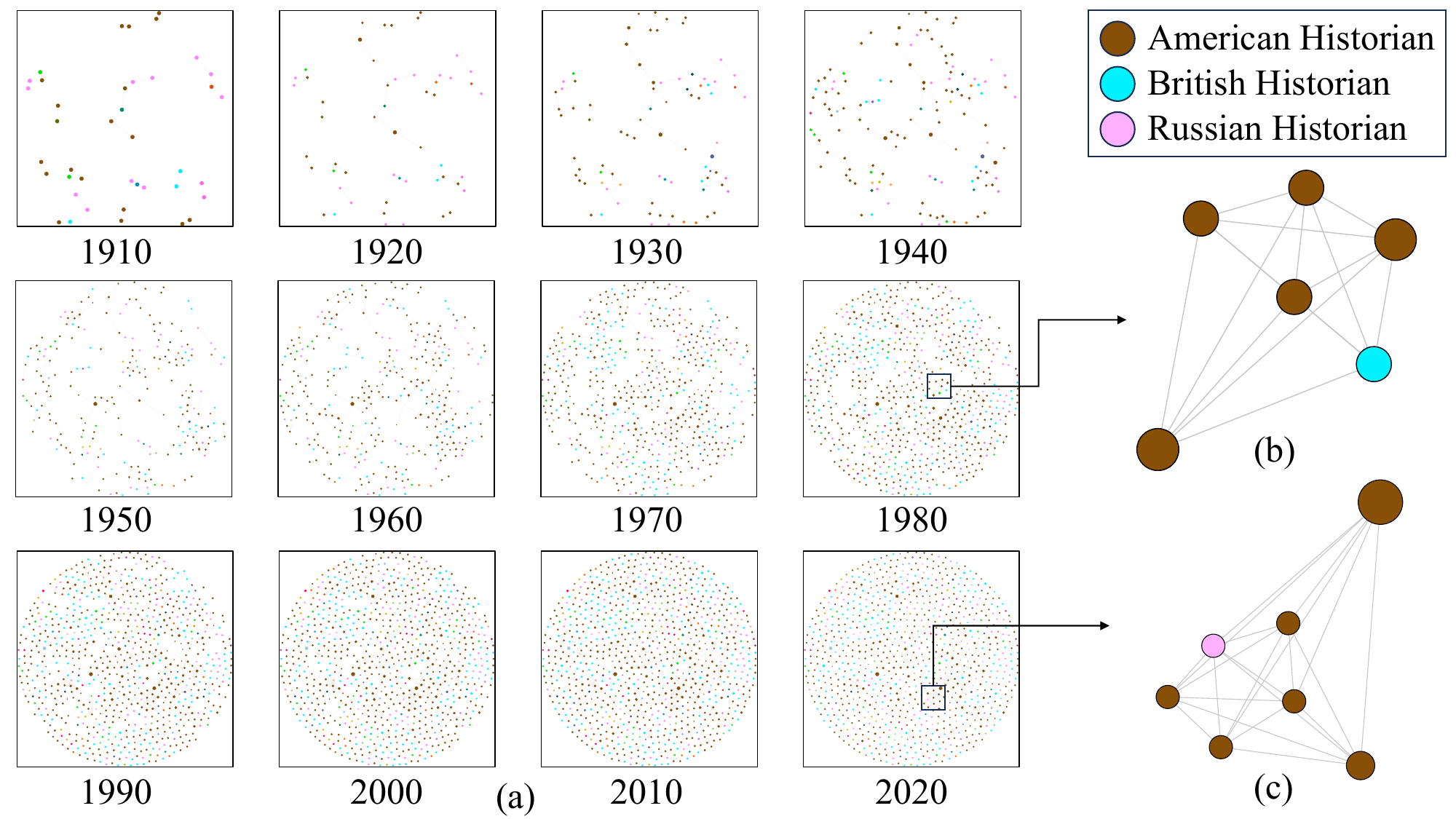}
    \caption{
    Dynamic interaction network comprising 899 historians.
    (a) Snapshots of the network every 10 years from 1910 to 2020. Nodes represent historians, the sizes of nodes are the PageRank values, and their nationalities are indicated by colors. The visualization is created using the Fruchterman Reingold layout.
    (b) and (c) zoom in on two connected components in the 1980 snapshot and 2020 snapshot respectively.
    }
    \label{fig:interactions}
\end{figure}

\subsubsection{Spatio-Temporal Interaction Network}
We construct the interaction network based on the temporal and spatial intersections of historians' trajectories, which can not be easily created from other datasets. 
We restrict the location must contain words related to schools and institutes such as University, College, etc. to ensure that the interactions are most likely to have happened, instead of merely capturing coarse spatio-temporal co-occurrences.
Overall, the network contains 899 nodes (historians) and 791 edges (interactions), spanning from 1811 to 2019. In Figure~\ref{fig:interactions}, we see graph snapshots taken every 10 years from 1910 to 2020.
Node sizes are calculated based on PageRank and the colors indicate the nationalities of the historians. It seems that from 1940 to 1970, there is a rapid increase in nodes and edges, compared with other periods of time. As a comparison, from 2000 to 2020, the increase appears to have slowed down. 
These are further supported by the distribution of historians' birth years, which are concentrated around 1920-1940.
We suspect that similar dynamics may also be reflected in the publication/citation records of history papers. Subplots (a) and (b) demonstrate two connected components from a microscopic point of view and from the colors we can see that these interactions mostly happen between historians from the same countries.

Beyond these analytics, the popular Spatio-Temporal Graph Convolutional Networks~\cite{yu2017spatio, Hu2023STGCN, lian2023ptp} may help better model the graph and provide predictive insights.

As an anecdote, \textit{William H. McNeill} from the University of Chicago, who died at the age of 98, had the highest PageRank score in this interaction graph.

\section{Conclusion and Future Work}
We propose a new task of extracting life trajectories from Wikipedia and introduce COSMOS, to effectively extract life trajectories from Wikipedia biography pages by combining the idea of contrastive learning and semi-supervised learning. To validate the method and showcase the potential of the extracted data, we extract the trajectories of historians, and perform an analysis based on the resulting trajectories.
We hope the open-sourced code, the million-level extracted trajectories, and the \textit{WikiLifeTrajectory} ground truth dataset, can support the trajectory extraction research and the analytical studies based on these trajectories.
As a largest of its kind, our dataset can be the basis for data-driven grand narratives and explorations of human mobility and interaction mechanisms. Beyond births and deaths, this comprehensive compilation encompasses various life milestones, offering insights into aspects such as education, work, and marriage. 
All our data shared from this work will be made FAIR~\cite{fair}.

We have to note that since we choose to extract trajectories from the English Wikipedia, there can be a bias that the extracted people are more likely to be from the English world~\cite{Roy2021InformationAI}. This should be considered when any research tries to draw conclusions from our dataset. To mitigate this, a possible future step is to extend our framework to versions of Wikipedia in other languages and further explore different designs of extraction algorithms. As another future improvement, we may include identifying triplet types such as ``graduate study'', ``attending a conference'', and ``delivering a speech'' in our task, to extract the purposes of trajectories.

\section{Ethics Statement}

Our study uses the English Wikipedia, and we ensure that our data collection process does not violate any privacy or confidentiality concerns.
A potential ethical concern is the misuse to extract the life trajectories of individual users (non-famous people). However, since our framework relies on a detailed description of one's life, the risk would arise from the leakage of personal information in such a description, rather than the framework itself.
Therefore, we believe that there are no essential ethical questions raised by our study.

\bibliography{main}

%%%%%%%%%%%%%%%%%%%%%%%%%%%%%%%%%%%%%%%%%%%%%%%%%%%%%%%%%%%%%%%
% Paper CheckList
\clearpage
\section{Paper Checklist}

\begin{enumerate}
\item For most authors...
\begin{enumerate}
    \item  Would answering this research question advance science without violating social contracts, such as violating privacy norms, perpetuating unfair profiling, exacerbating the socio-economic divide, or implying disrespect to societies or cultures?
    \answerYes{Yes, see the Introduction, Ethics Statement, Conclusion and Future Work.}
  \item Do your main claims in the abstract and introduction accurately reflect the paper's contributions and scope?
    \answerYes{Yes, see the Abstract and Introduction.}
   \item Do you clarify how the proposed methodological approach is appropriate for the claims made? 
    \answerYes{Yes, see the Introduction, Experimental Results, Conclusion and Future Work.}
   \item Do you clarify what are possible artifacts in the data used, given population-specific distributions?
    \answerYes{Yes, see the  Conclusion and Future Work.}
  \item Did you describe the limitations of your work?
    \answerYes{Yes, see the Conclusion and Future Work.}
  \item Did you discuss any potential negative societal impacts of your work?
    \answerYes{Yes, see the Ethics Statement.}
    \item Did you discuss any potential misuse of your work?
    \answerYes{Yes, see the Ethics Statement.}
    \item Did you describe steps taken to prevent or mitigate potential negative outcomes of the research, such as data and model documentation, data anonymization, responsible release, access control, and the reproducibility of findings?
    \answerYes{Yes, see the Ethics Statement. The data we use is from Wikipedia, a publicly available website.}
  \item Have you read the ethics review guidelines and ensured that your paper conforms to them?
    \answerYes{Yes.}
\end{enumerate}

\item Additionally, if your study involves hypotheses testing...
\begin{enumerate}
  \item Did you clearly state the assumptions underlying all theoretical results?
    \answerNA{NA}
  \item Have you provided justifications for all theoretical results?
    \answerNA{NA}
  \item Did you discuss competing hypotheses or theories that might challenge or complement your theoretical results?
    \answerNA{NA}
  \item Have you considered alternative mechanisms or explanations that might account for the same outcomes observed in your study?
    \answerYes{Yes, see the Experimental Results.}
  \item Did you address potential biases or limitations in your theoretical framework?
    \answerYes{Yes, see the Dataset. We use stratified sampling according to occupation to avoid potential bias.}
  \item Have you related your theoretical results to the existing literature in social science?
    \answerYes{Yes, see the Analysis of a Sample Set, Conclusion and Future Work.}
  \item Did you discuss the implications of your theoretical results for policy, practice, or further research in the social science domain?
    \answerYes{Yes, see the Introduction, Analysis of a Sample Set, Conclusion and Future Work.}
\end{enumerate}

\item Additionally, if you are including theoretical proofs...
\begin{enumerate}
  \item Did you state the full set of assumptions of all theoretical results?
    \answerNA{NA}
	\item Did you include complete proofs of all theoretical results?
    \answerNA{NA}
\end{enumerate}

\item Additionally, if you ran machine learning experiments...
\begin{enumerate}
  \item Did you include the code, data, and instructions needed to reproduce the main experimental results (either in the supplemental material or as a URL)?
    \answerYes{Yes, see the open-sourced repository \url{https://anonymous.4open.science/r/wiki_life_trajectory/}.}
  \item Did you specify all the training details (e.g., data splits, hyperparameters, how they were chosen)?
    \answerYes{Yes, see the Experiments.}
     \item Did you report error bars (e.g., with respect to the random seed after running experiments multiple times)?
    \answerNo{No, because our experiment is conducted using only one fixed random seed of 42 and controls all random numbers to ensure that our experimental results can be reproduced.}
	\item Did you include the total amount of compute and the type of resources used (e.g., type of GPUs, internal cluster, or cloud provider)?
    \answerYes{Yes, see the Implementation Details section in Experiments.}
     \item Do you justify how the proposed evaluation is sufficient and appropriate to the claims made? 
    \answerYes{Yes, see the Experimental Results section in Experiments.}
     \item Do you discuss what is ``the cost`` of misclassification and fault (in)tolerance?
    \answerYes{Yes, see the Experimental Results section in Experiments.}
  
\end{enumerate}

\item Additionally, if you are using existing assets (e.g., code, data, models) or curating/releasing new assets, \textbf{without compromising anonymity}...
\begin{enumerate}
  \item If your work uses existing assets, did you cite the creators?
    \answerYes{Yes, see the Method.}
  \item Did you mention the license of the assets?
    \answerNo{No, because all the tools, algorithms and data we use are publicly available.}
  \item Did you include any new assets in the supplemental material or as a URL?
    \answerYes{Yes, see the URLs in footnotes.}
  \item Did you discuss whether and how consent was obtained from people whose data you're using/curating?
  \answerYes{Yes, see the Ethics Statement. The data we use is from Wikipedia, a publicly available website.}
  \item Did you discuss whether the data you are using/curating contains personally identifiable information or offensive content?
  \answerYes{Yes, see the Ethics Statement. The data we use is from Wikipedia, a publicly available website.}
    \item If you are curating or releasing new datasets, did you discuss how you intend to make your datasets FAIR (see \citet{fair})?
    \answerYes{Yes, see the Conclusion and Future Work.}
    \item If you are curating or releasing new datasets, did you create a Datasheet for the Dataset (see \citet{gebru2021datasheets})? 
    \answerYes{Yes, we follow the instructions and create a Datasheet.}
\end{enumerate}

\item Additionally, if you used crowdsourcing or conducted research with human subjects, \textbf{without compromising anonymity}...
\begin{enumerate}
  \item Did you include the full text of instructions given to participants and screenshots?
    \answerNA{NA}
  \item Did you describe any potential participant risks, with mentions of Institutional Review Board (IRB) approvals?
    \answerNA{NA}
  \item Did you include the estimated hourly wage paid to participants and the total amount spent on participant compensation?
    \answerNA{NA}
   \item Did you discuss how data is stored, shared, and deidentified?
   \answerNA{NA}
\end{enumerate}
\end{enumerate}
%%%%%%%%%%%%%%%%%%%%%%%%%%%%%%%%%%%%%%%%%%%%%%%%%%%%%%%%%%%%%%%
\clearpage

\appendix
\section{Appendix}
\subsection{Prompt for GPT-3.5}
\subsubsection{Prompt for Annotating Triplets}

Figure~\ref{fig:apd1} and~\ref{fig:apd2} show the given prompts when we use GPT-3.5 (gpt-3.5-turbo-0613) for data annotation. For each input sentence, we first extract triplets through the prompt of Figure~\ref{fig:apd1}, and then use the prompt of Figure~\ref{fig:apd2} to perform self-verification. During the annotation step, we set the temperature (the parameter used to adjust the diversity of the model's output) to 0.8 and three trials are performed for each input. In the self-verification step, the temperature is set to 0 since we need GPT-3.5 to give stable final results. 

\subsubsection{Prompt for the Baseline Method}
Figure~\ref{fig:apd3} shows the prompt provided when we use GPT-3.5 (gpt-3.5-turbo-0613) as the baseline. The temperature here is set to 0 and one trial is performed for each input.

\subsection{Performance of GPT-4}
\begin{figure}[!htbp]
    \centering
    \subfigure[]{
    \begin{minipage}[!htb]{1\linewidth}
        \centering
        \includegraphics[width=1\textwidth]{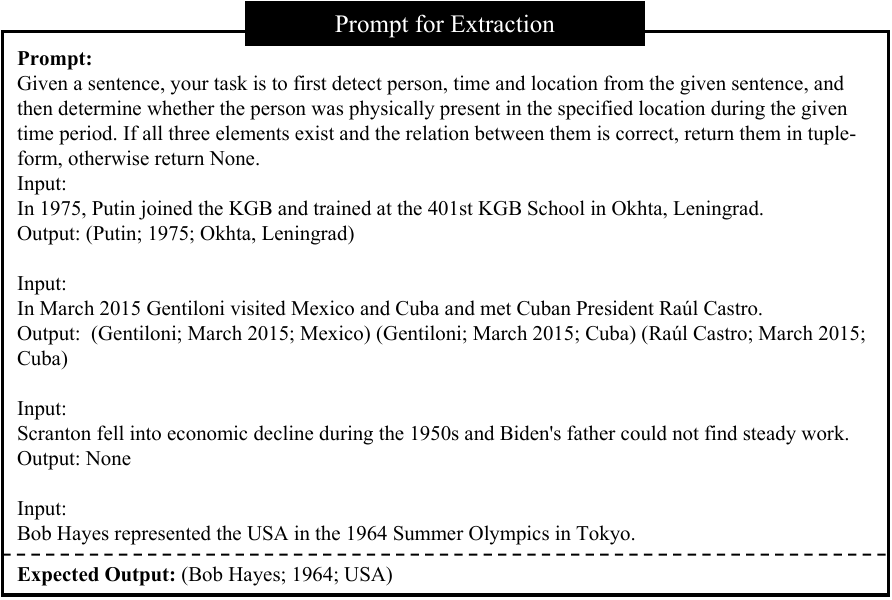}
        \label{fig:apd1}
    \end{minipage}%
    }%
    \quad
    \subfigure[]{
    \begin{minipage}[!htb]{1\linewidth}
        \centering
        \includegraphics[width=1\textwidth]{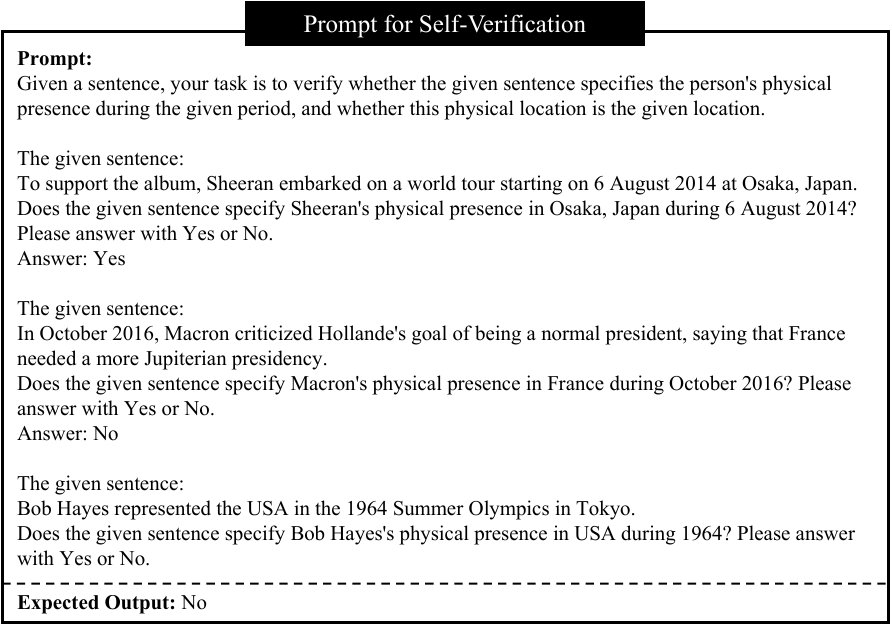}
        \label{fig:apd2}
        \end{minipage}%
    }
    \quad
    \subfigure[]{
    \begin{minipage}[!htb]{1\linewidth}
        \centering
        \includegraphics[width=1\textwidth]{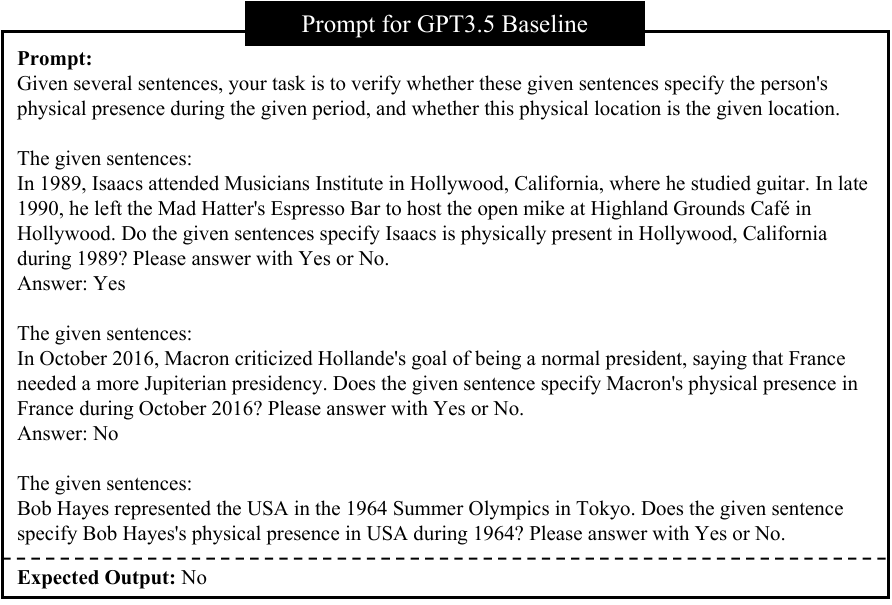}
        \label{fig:apd3}
    \end{minipage}%
    }%
    \centering
    \caption{Prompt for GPT.}
    \label{fig:propmt4GPT}
\end{figure}

To evaluate the performance of GPT-4 on the life trajectory extraction task, we test it on a quarter of the test set (643 instances random sampled from the ``Representative'' test set). During the experiments, we set the temperature to 0 and use the same prompt given to GPT-3.5 (see Figure~\ref{fig:apd3}), with one trial for each input, which costs about \$5 in total (around 250,000 tokens).

The experimental results are reported in Table~\ref{tab: gpt4 results}.
It is observed that GPT-4 outperforms GPT-3.5 in overall performance (increased by 8.33\% in F1), while it is still weaker than most supervised methods. As general-purpose language models, ChatGPT-like LLMs might not yet be ready to be directly applied in large-scale extraction tasks instead of specialized models~\cite{foppiano2024mining}.

%%%%%%%%%%%%%%%%%% gpt4 baseline Results   %%%%%%%%%%%%%
\begin{table}[htb]
\scriptsize
\setlength\tabcolsep{1pt}
  \centering
    \begin{tabular}{cccccccccc}
    \toprule
    % \midrule
    \multirow{3}{*}{} & \multicolumn{4}{c}{\textbf{Representative}} & \multicolumn{4}{c}{\textbf{Representative$_\textrm{m}$}} & \multicolumn{1}{c}{\textbf{Representative$_\textrm{g}$}} \\
    % subtitle
    \cmidrule(lr){2-5} \cmidrule(lr){6-9} \cmidrule(lr){10-10}
    & Acc (\%) & P (\%) & R (\%) & F1 (\%) & Acc (\%) & P (\%) & R (\%) & F1 (\%) & R (\%)\\
    \midrule
    GPT-4  & 79.63 & 75.53 & 83.33 & 79.24 & 76.56 & 61.61 & 76.92 & 68.42 & 91.60\\
\bottomrule
    \end{tabular}
    \caption{Results of GPT-4 baseline. From the ``Representative'' test set, 643 samples are randomly sampled (512 manually labeled, 131 GPT-3.5 labeled).}
    \label{tab: gpt4 results}%
\end{table}%
%%%%%%%%%%%%%%%%%% gpt4 baseline Results   %%%%%%%%%%%%%

Additionally, we notice the Recall of GPT-4 is much lower than that of GPT-3.5 (decreased by 11.79\%), which is also observed by other researchers when extracting biological information~\cite{BABAIHA2024100095}. Why GPT-based models produce such a phenomenon may be of interest to LLMs researchers.

% \nolinenumbers
\end{document}